\pdfoutput=1

\documentclass[11pt]{article}

\usepackage[preprint]{acl}

\usepackage{times}
\usepackage{latexsym}

\usepackage[T1]{fontenc}

\usepackage[utf8]{inputenc}

\usepackage{microtype}

\usepackage{inconsolata}

\usepackage{graphicx}

\usepackage{enumitem}
\usepackage{algorithm}
\usepackage{tabularx}
\usepackage{subcaption}
\usepackage[most]{tcolorbox}
\definecolor{tablerow1}{RGB}{225,217,205}
\definecolor{tablerow2}{RGB}{236,229,221}
\usepackage{algpseudocode}
\usepackage{bbm}
\usepackage{amsmath}
\usepackage{multirow}
\usepackage{booktabs}
\usepackage{url}
\usepackage{graphicx}
\definecolor{RoseQuartzBg}{HTML}{F7CAC9}
\definecolor{RoseQuartz}{HTML}{F5A798}
\definecolor{Serenity}{HTML}{92A8D1}
\definecolor{OrangeRed}{rgb}{1.0, 0.27, 0.0}
\definecolor{Red}{rgb}{1.0, 0.0, 0.0}
\definecolor{Turquoise}{HTML}{0F4C81}
\usepackage{xparse}
\usepackage{soul}
\NewDocumentCommand{\lifu}{ mO{} }{\textcolor{OrangeRed}{\textsuperscript{\textit{Lifu}}\textsf{\textbf{\small[#1]}}}}
\NewDocumentCommand{\sijia}{ mO{} }{\textcolor{blue}{\textsuperscript{\textit{Sijia}}\textsf{\textbf{\small[#1]}}}}
\usepackage{amssymb}
\usepackage{array}
\newcolumntype{L}{>{\centering\arraybackslash}m{4cm}}
\newcolumntype{M}{>{\centering\arraybackslash}m{3cm}}
\newcolumntype{S}{>{\centering\arraybackslash}m{2cm}}
\newcolumntype{P}{>{\arraybackslash}m{12cm}}
\newcolumntype{Q}{>{\arraybackslash}m{6cm}}
\usepackage{xspace}
\usepackage{xcolor}
\usepackage{ulem}
\usepackage{mdframed}
\usepackage{fontawesome}
\definecolor{ao}{rgb}{0.0, 0.5, 0.0}
\definecolor{forestgreen}{rgb}{0.13, 0.55, 0.13}
\newcommand{\yesmark}{\color{forestgreen}{\faCheckSquare}}
\newcommand{\nomark}{\color{red}{\faTimesCircle}}
\usepackage{xspace}
\newcommand{\model}{\textbf{\textsc{Debate-EE}}\xspace}

\title{Debate as Optimization: Adaptive Conformal Prediction and Diverse Retrieval for Event Extraction}

\author{Sijia Wang,  \ Lifu Huang
\\
  Virginia Tech
 \\
  {\tt \{sijiawang,lifuh\}@vt.edu}
  }

\begin{document}
\maketitle
\begin{abstract}

We propose a multi-agent debate as optimization (DAO) system for event extraction, where the primary objective is to iteratively refine the large language models (LLMs) outputs through debating without parameter tuning.
In DAO, we introduce two novel modules: the Diverse-RAG (DRAG) module and the Adaptive Conformal Prediction (AdaCP) module. DRAG systematically retrieves supporting information that best fits the debate discussion, while AdaCP enhances the accuracy and reliability of event extraction by effectively rejecting less promising answers.
Experimental results demonstrate a significant reduction in the performance gap between supervised approaches and tuning-free LLM-based methods by 18.1\% and 17.8\% on ACE05 and 17.9\% and 15.2\% on CASIE for event detection and argument extraction respectively.
\end{abstract}

\section{Introduction}

Event extraction (EE)~\cite{grishman1997information,chinchor1998muc,ahn2006stages} involves identifying and categorizing event mentions, expressed through trigger tokens and participants in natural language text. 
Recent studies show that leveraging Large Language Models (LLMs) has led to remarkable advancements in numerous applications ~\cite{touvron2023llama, zhang2022opt, anil2023palm, openai_gpt3, openai_gpt4}. 
Their potent natural language understanding capabilities are generic and adaptable to nearly any open domain.
However, a significant gap remains for event extraction between advanced tuning-based approaches ~\cite{wadden-etal-2019-entity, yinglinACL2020, naacl2022degree, xinyaduEMNLP2020, WangAcl2022_query, zhao2023demosg} and approaches without tuning ~\cite{li2023evaluating,han2023information, wei2024chatie}.

LLMs struggle to match the performance of tuning-based approaches due to several challenges.
First, the inherent ambiguities and variations in event mentions present significant obstacles in accurately identifying them.
For instance, in the phrase ``pay the fines'', two potential questions arise: whether the event type should be classified as a \texttt{Transfer-Money} or \texttt{Fine} event and whether the event trigger should be ``pay'' or ``fines''. 
Second, existing solutions fail to efficiently incorporate domain-specific knowledge, such as extensive event schemas. While a common solution is to enumerate event schemas into the prompt \cite{lin-etal-2023-global, wang-etal-2023-code4struct}, LLMs can struggle to fully comprehend and utilize this information. 
Lastly, unlike tuning-based methods that can leverage annotated data, such as ACE05 \cite{ldc_ace05} and ERE \cite{song2015light}, to learn implicit statistical features and resolve nuanced semantic differences, LLMs are difficult to tune, even with small amounts of data, particularly without access to the model checkpoint.

To address these challenges, we introduce a tuning-free multi-agent Debating-as-Optimization (DAO) framework. This approach demonstrates that event extraction answers can be gradually optimized through debates among LLM agents without domain-specific fine-tuning, allowing the system to adapt effortlessly to new domains or ontologies. To optimize the initial solution, we propose two novel modules: the diverse retrieval augmented module (DRAG) and the adaptive conformal prediction module (AdaCP). The DRAG module dynamically retrieves domain-specific data entries that best fit the current points of disagreement. The AdaCP model employs an adaptive conformal prediction policy to progressively reject less convincing answers based on the retrieved knowledge. The event extraction answer is gradually refined through more precise retrieval of domain-specific knowledge and the application of stricter rejection rules. Our aim is to demonstrate that the significant performance gap can be narrowed with the proposed multi-agent debate framework.

The contribution of the proposed work includes
\begin{itemize}[noitemsep,nolistsep,wide,itemindent=10pt]
\item A novel multi-agent debate framework is introduced, which highlights the refining of event extraction answers through a debating process.
\item An Adaptive Conformal Prediction module, AdaCP, is proposed to systematically reject less convincing answers. 
\item A Diverse-RAG Module (DRAG) is developed, featuring dynamic clustering techniques to accurately retrieve reference information crucial for achieving correct outcomes.
\item Though the performance gap against fine-tuning-based approaches persists, significant improvements are achieved across various datasets.

\end{itemize}

\section{Related Work}

\paragraph{LLMs for Event Extraction}
Early studies \cite{gao2023exploringChatGPTee, li2023evaluating, wei2024chatie, han2023information} utilized specific guidelines or instructions to prompt the LLMs to directly perform inference on event extraction.
However, the experimental results reveal that current LLMs may lack the comprehensive event schema knowledge necessary for extracting event information effectively from text.
Recent investigations~\cite{lin-etal-2023-global, han2023information, guo2023retrieval} have delved into in-context learning, wherein task instructions and a few in-context examples are provided. However, their empirical results highlight a significant performance disparity between in-context learning and approaches relying on fine-tuning.

\paragraph{Multi-agent System}
Multi-agent collaboration has drawn considerable attention benefit from the development of autonomous agents based on LLMs, including GPTs~\cite{brown2020language, openai_gpt3, openai_chatgpt, openai_gpt4}, Antrophic LMs, LLaMAs~\cite{touvron2023llama, touvron2023llama2}, PaLM~\cite{chowdhery2022palm, anil2023palm}, etc.. There are two categories of interactions for multi-agent systems, cooperative interaction and adversarial interaction. 
Agents in cooperative interaction are carefully designed to serve their duties and work together to finish the task \cite{zhou2023agents, wu2023autogen, park2023generative, qian2023communicative, chen2023agentverse}. 
On the other hand, adversarial interactive approaches are designed to derive accurate and consistent conclusions in a debating manner. Adversarial multi-agent debate systems mostly consist of multiple debaters~\cite{du2023improving}, with the choice to intergrate a summarizer~\cite{chan2023chateval}, a judge~\cite{liang2023encouraging}, and a critic agent~\cite{fu2023improving, wang2023apollos}. 
The challenge in implementing a multi-agent debate system for information extraction lies in determining how to retrieve essential information and steer the discussion effectively.

\paragraph{Retrieval Augmented Generation}

Retrieval Augmented Generation (RAG) has proven to be effective across various recent applications \cite{10.5555/3495724.3496517, glass-etal-2022-re2g, chen-etal-2022-murag, siriwardhana-etal-2023-improving, 0011LH024}. Existing RAG methods proposed advanced strategies concerning \textit{what to retrieve} and \textit{when to trust} the retrieved content. For example, \cite{li-etal-2023-web} and \cite{jiang-etal-2023-active} advocate for retrieval based on the confidence level of the LLMs regarding the content. 
\cite{zhang2023repocoder} propose a method for progressively retrieving relevant code snippets in code completion. 
\citet{asai2024selfrag} and \citet{wu2024repoformer} suggest selecting retrieved content depending on output quality, leveraging the self-reflection and self-evaluation capabilities of the LM. 
However, the exploration of progressively retrieving more fine-grained content to benefit complex inquiries remains relatively unexplored. This work takes one step forward by advocating retrieval with conformal prediction and adaptively retrieving more fine-grained content, consequently enhancing decision-making processes.

w
\section{Approach}

\begin{figure*}[t]
  \centering
  \includegraphics[width=\textwidth]{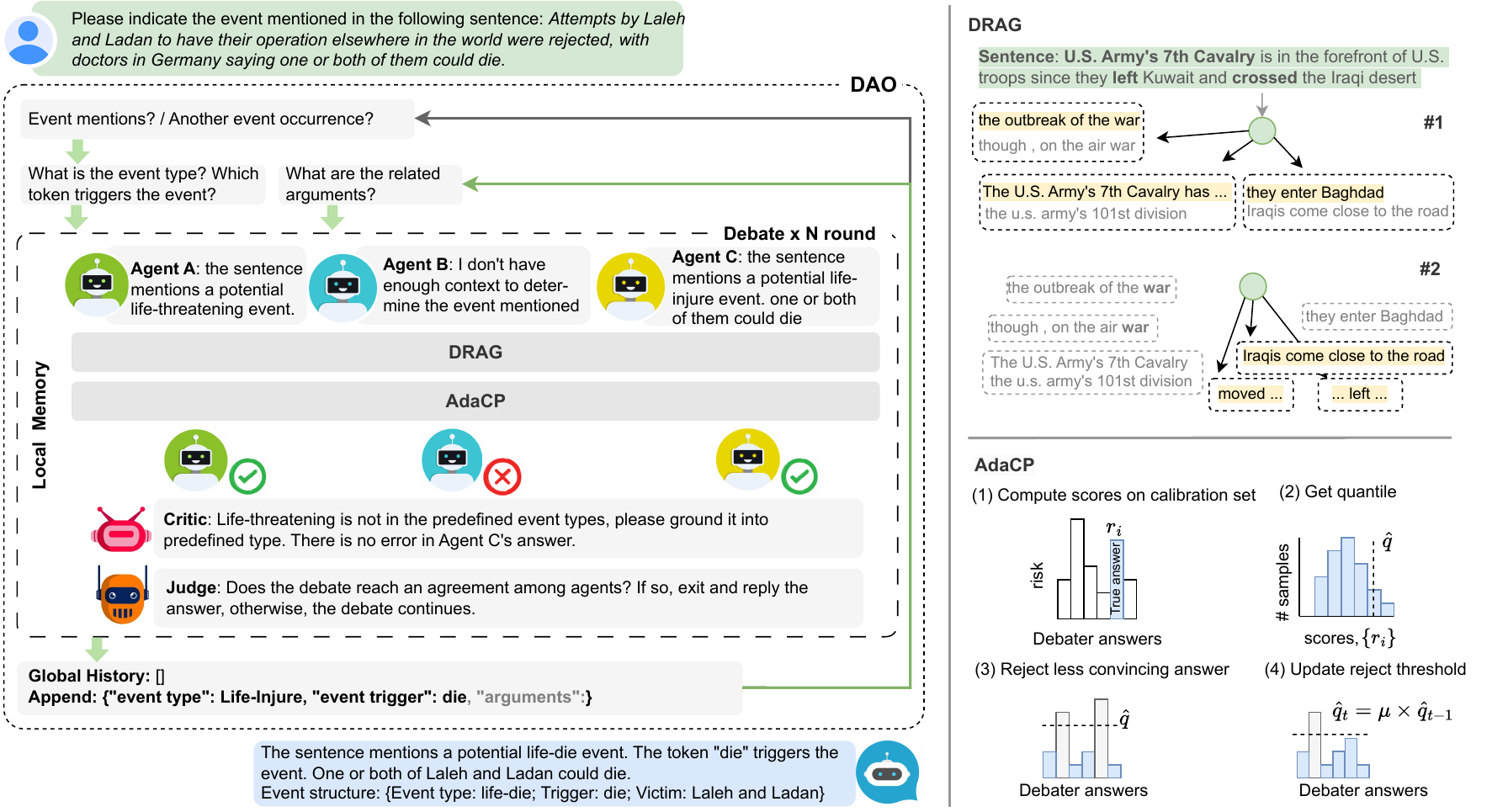}
  \caption{Debate As Optimization (DAO) framework}
  \vspace{-0.5cm}
  \label{fig:debate_framework}
\end{figure*}

\newtcolorbox{mybox}[1]{colback=tablerow1!5!white,
colframe=tablerow1!75!black,fonttitle=\bfseries,
title={#1}, left=2mm, right=2mm}
\setlength{\abovedisplayskip}{0.5pt}
\setlength{\belowdisplayskip}{0.5pt}
In event extraction (EE), two sub-tasks are involved: event detection (ED) and event argument extraction (EAE). The proposed Debating as Optimization (DAO) framework tackles both ED and EAE through a unified debating process, employing distinct task-specific prompts for each sub-task. Detailed agent prompts are in Appendix \ref{sec:prompts}.

\subsection{Problem Formulation} 
The task of EE is to identify event mentions within a sentence, which consist of an event trigger and related event arguments. In formal terms, given a sentence $w=\{w_1, ..., w_n\}$ and a specified target event type $e_i$, an EE system aims to extract the event trigger $t$ and its associated argument mentions $a=\{a_1, ..., a_g\}$. 
In this work, we focus on in-context learning (ICL) with $M$ sample selection, where $M$ indicates the maximum number of examples to be included in the system. Formally, in-context learning with $M$ sample selection can be outlined as follows: given a sentence $w$, a dataset $\mathcal{D}$, a set of $M$ examples $\mathcal{D}(M)=\{d_1, ..., d_m| m\leq M\}$ can be sampled as in-context examples for inference on each $w$.
This is an instance-based in-context example selection setting designed to exploit the event extraction capabilities and reasoning capabilities of LLMs with limited computation and without tuning.

\subsection{Debate as Optimization}
\subsubsection{Debate Agents}
As shown in Figure \ref{fig:debate_framework}, the proposed debate framework consists of four types of agents: the Debaters, the Critic, the Judge, and the Summarizer. 
Each debating agent role is designed to serve specific responsibilities to optimize the final solution. 
\textbf{Debaters} are the agents that generate opinions and defend or adjust opinions based on the given information. Given a specific question, the debaters first need to generate preferably different opinions. Depending on the retrieved information, the debaters will also reason, defend, or adjust their solution.
The \textbf{Critic} is asked to identify any potential errors that have been made by the debaters.
The responsibility of the \textbf{Judge} is to determine whether the debaters have reached an agreement on their solution. 
The \textbf{Summarizer} collects all the pieces of commonly agreed solutions and formalizes the final solution. 

\subsubsection{Multi-Agent Debate Process}
A single round of the debating process consists of four stages: Initial Opinion Rendering, Event Information Retrieval, Cross-Examination, and Judgement. 
During the \textbf{Initial Opinion Rendering} stage, we aim to collect diverse opinions from the debaters. This diversity can be achieved by setting different temperatures or leveraging different LLMs, such as using ChatGPT and Gemini as debaters. The prompt for this stage is outlined as follows:
\begin{mybox}{Debater Prompt}
Given sentence: **[SENT]** Answer the following question: [TASK\_INSTRUCTION]
\end{mybox}
\noindent It is essential that responses are as accurate as possible; thus, detailed task instructions are preferred.

Next, we retrieve two categories of event information for the \textbf{Event Information Retrieval} stage: (1) The event definition and descriptions from the event extraction guideline for every event type mentioned in the initial opinions, and (2) Examples retrieved by the proposed retrieval module {(details are in Section \ref{sec:AdaCP})}. The acquired knowledge will then be broadcast to all the debating agents, excluding the Judge, since the Judge's decisions should be solely based on the consensus reached, rather than the specific content of the discussion. 

Every opinion rendered together with all the retrieved event information will be validated by an adaptive conformal prediction module, AdaCP, which is described in Section \ref{sec:AdaCP}.
Agents whose opinions have successfully passed AdaCP will proceed to the \textbf{Cross-Examination} (CE) stage. 
This process comprises two components: debaters engage in debates with each other, while the Critic agent identifies potential flaws in the debaters' responses. 
The prompt for the debaters in this stage is as follows: 
\begin{mybox}{Debater CE Prompt}
Carefully review the information in the event definitions and retrieved examples. Defend your answer, or update your answer. 
\end{mybox}
\noindent The prompt for the Critic agent is designed to be more informative. Our preliminary studies show that it it beneficial to include some common mistakes in event extraction would be helpful. For example, the CE prompt for the Critic in ED is as follows:
\begin{mybox}{Critic CE Prompt}
After reviewing the event definition and examples, assess whether the identified event type and event trigger align with the event occurrence in the sentence. Consider whether there is any other event type that better matches the event mentioned in the sentence. Respond succinctly with your judgment.
\end{mybox}
At the end of each round of debate, we ask the Judge agent to make a \textbf{Judgement} on whether we have reached a consensus on the debate topic or if further debate is required. For example, the judge prompt for ED is as follows: 
\begin{mybox}{Judge Prompt}
Do debaters and the critic reach an agreement on event type and trigger extraction? If so, reply in a table. The header of the table is | event type | event trigger |. If disagree, require reply: **No agreement, debate continues**. If both debaters believe there is no event mention involved, reply **No event**.
\end{mybox}
\noindent A round of debate concludes either when the maximum number of rounds is reached or when the judge decides an agreement has been reached. If an event type and event trigger are identified during the ED procedure, the system proceeds to debate argument extraction. Otherwise, it skips argument extraction.

\subsubsection{Diverse-RAG}
\label{sec:drag}
The Diverse-RAG (DRAG) module dynamically retrieves event related data entries that best fit the current points of disagreement. 
It is crafted around four key principles:
(1) \textbf{Distance}. To enhance the informativeness of retrieved examples, we prioritize semantic proximity. Utilizing a sentence encoding method $\mathrm{emb}(\cdot)$, we encode both the input context $\mathrm{x}$ and reference texts $\mathrm{Y}=\{\mathrm{y_j}\}_{j=0}^{N_{ref}}$
\begin{align*}
\mathbf{x} = \mathrm{emb}(\mathrm{x}), \mathbf{Y} = \{\mathrm{emb}(\mathrm{y_j})\}_{j=0}^{N_{ref}}
\end{align*}
The retrieval module then selects the top-K sentences closest in semantic representation. In our experiments, we set K to 128.
(2) \textbf{Diversity}. Within the Top-K retrieved reference texts, some examples may share common information that is not necessarily pertinent to the target event. For instance, identical long entity spans can inflate similarity scores. To address this, we employ clustering to group similar examples, mitigating redundancy. The clustering operation can be expressed as
\begin{align*}
\min \sum_{j=1}^{K} \mathrm{dis}(c_p, y_j)^2 \\
s.t. \;\;\mathrm{dis}(c_{p_i}, c_{p_j}) > \mu
\end{align*}
where $\mu$ is the clustering threshold. Exclusively one data entry from each cluster can be selected to be included in reference sentences for the current round. Additionally, the closest $M$ data points from $M$ distinct clusters are selected as the final reference data entries. 
(3) \textbf{Polarity}. Effective event extraction requires consideration of both positive and negative reference event mentions. For instance, a token like "meeting" may or may not trigger a specific event category. Therefore, both positive and negative event mentions are included in the retrieval. 
(4) \textbf{Adaption}. We conceptualize debating as an optimization process, evolving from broad to fine-grained retrieval. Initially, retrieval aims for breadth, gradually transitioning to more refined searches as the debate progresses. This evolution is captured through the decay of cluster radius over time, which can be formally expressed as 
\begin{align*}
\mu_t = \lambda * \mu_{t-1}
\end{align*}
where $\mu_{t-1}$ is the clustering radius of the previous round, and $\lambda$ is the cluster radius decay factor.

\subsubsection{Adaptive Conformal Prediction}
\label{sec:AdaCP}
The objective of Adaptive Conformal Prediction (AdaCP) is to progressively reject less convincing answers.
Previous conformal prediction techniques \cite{10.5555/1390681.1390693, 10.5555/2074094.2074112,  10.5555/1062391, 10.1080/01621459.2012.751873, 10.1145/3478535, angelopoulos2022learn, yang2024selection,  quach2024conformal} generate a range of predictions encompassing the true output with a predetermined level of confidence. 
Our framework goes beyond the standard by actively updating the conformal calibration configuration, iteratively rejecting less convincing answers based on the retrieved knowledge.

Formally, conformal prediction either accepts or rejects the null hypothesis that the pairing $(x, y)$ is correct. The test method is a nonconformity measure, $R((x, y), \mathcal{D})$, where $\mathcal{D}$ is a calibration dataset with annotated examples. Intuitively, a lower value of $R$ reflects that point $(x, y)$ ``conforms'' to $D$, whereas a higher value of $M$ reflects that $(x, y)$ does not. 
Consider a calibration set $\mathcal{D}_{cal} = \{(x_i, y_i)\}_{i=1}^{N_{cal}}$, where $N_{cal}$ is the calibration set size. The conformal generation risk is set as the $1-\delta$ quantile of the risk scores
\begin{align*}
\hat{q}_0 =\mathrm{Quantile} (\{r_1, \cdot, r_n\}, \frac{\lceil(n+1)(1-\delta)\rceil}{n}),
\end{align*}
where $r_i=R(x_i, y_i)$, and $R(x, y):\mathcal{X}\times\mathcal{Y}\rightarrow\mathbb{R}$ is an independent quality function, such as using the negative log-likelihood function of a standalone LM. The assumption is that for a fair-quality LM, the likelihood of a correct answer has a higher probability. 
The coverage guarantee confirms that the prediction set after calibration contains the true answer at risk level $\delta$, which can be denoted as  
$
    \mathbb{P}[R(x, y)\le \hat{q}] \ge 1-\delta.
$
At inference time, we reject a debater's answer if $R(x, y)> \hat{q}$.

Additionally, given the debating design of our system with retrieval, the conversation continues with increasing content and information. Then the risk score can be updated as $r_i=R(x_i\oplus c, y_i)$, where $c$ denotes the retrieved information. The risk score is expected to decrease with properly retrieved information. 
Thus we propose an adaptive nonconformity measure with a constant decay rate 
\begin{align*}
\hat{q}_t = \beta \times \hat{q}_{t-1} 
\end{align*}
where $\hat{q}_{t-1}$ is the nonconformity threshold of the debate round $t-1$, and $\beta$ is the decay factor. 
Intuitively, AdaCP starts with a more inclusive rejection configuration at the beginning of the debate process, allowing a broad range of potential event extraction answers to be considered. As the debate progresses and more event information is retrieved, the calibration model becomes more confident in identifying the accurate event answer. Consequently, a stricter policy is applied, progressively rejecting less convincing answers.

\section{Experimental Setup}

\paragraph{Dataset and Evaluation Metrics} 
We conducted experiments on two public benchmark datasets, ACE05-E (Automatic Content Extraction, ACE05)\footnote{\url{https://catalog.ldc.upenn.edu/LDC2006T06}} and CASIE~\cite{Satyapanich2020CASIEEC}. For the ACE05, we reported evaluation results on the test set using the same test split as in \cite{yinglinACL2020}. For the CASIE, we used the same test split as in \citet{han2023information}.
The evaluation is focused on three sub-tasks: ED, EAE where the ground truth trigger is given, and EE where ED and EAE are performed jointly. We only report argument extraction performance for EE following previous work~\cite{han2023information, guo2023retrieval}.
For the ACE05 dataset, we followed previous work \cite{yinglinACL2020} and used the Exact Match F1 score for evaluating ED and the Argument Head F1 score for evaluating EAE and EE. For the CASIE dataset, we adhered to the evaluation standards established in previous studies \cite{Satyapanich2020CASIEEC, han2023information}, employing the \texttt{types} metric for all three sub-tasks.

\begin{table*}[ht]
\centering\scalebox{0.7}{
\begin{tabular}{l|c|c| c|c |c|c|c|c}
\toprule
\multirow{2}{*}{\textbf{Method}}&\multirow{2}{*}{\textbf{Ontology usage}} & \multirow{2}{*}{\textbf{Paradigm}} & \multicolumn{3}{c|}{\textbf{ACE05}}&\multicolumn{3}{c}{\textbf{CASIE}}\\
 \cmidrule{4-6}\cmidrule{7-9}
&&&ED & EAE & EE&ED & EAE & EE\\
\midrule
\textbf{DEGREE} \cite{hsu-etal-2022-degree}                 
            &\yesmark       & SFT   
            & 73.3 &73.5 & 55.8 &-&-&-\\
\textbf{InstructUIE} \cite{wang2023instructuie}                 
            &\yesmark       & SFT   
            & 77.1      & 72.9& -&-&-&-\\
\textbf{RexUIE}\cite{liu-etal-2023-rexuie}&\nomark&SFT 
            & 73.3      & - & 57.3 &73.0       & -& 63.9\\
\midrule
\textbf{ChatGPT-14}~\cite{li2023evaluating}     
            & \nomark      & ZS
            &17.1 &28.9 & 7.3 &-&-&-\\            
      
\textbf{ChatIE}~\cite{wei2024chatie}      
            & \nomark   & ZS
            &-          & 29.5  & - &-&-&-\\
            
\textbf{ChatGPT-IE}~\cite{han2023information}  
            & \nomark   & ICL-5
            & 27.3   & 31.6  & 13.8& 18.2      & 27.4 & 19.0\\
      
\textbf{G-PTLM}~\cite{lin-etal-2023-global}
            & \yesmark  & ZS
            & -         &31.2 & - &-&-&-\\
\textbf{\textsc{Code4Struct}} \cite{wang-etal-2023-code4struct} 
            & \yesmark  & ZS
            & -         & 37.8 & -&-&-&-\\                       
\textbf{Code4UIE}~\cite{guo2023retrieval}   
            & \yesmark  & ICL-10$^*$
            & 37.4      & 57.0 & 21.3& 28.7      & - & 30.8\\

\midrule
    \model (Gemini-GPT)          &\yesmark   &ICL-10$^*$ & 50.2   & \textbf{59.5}  & 30.6 & \textbf{41.8} &\textbf{59.3} & \textbf{40.5}\\
    \model (Llama3-GPT)          &\yesmark   &ICL-10$^*$ & \textbf{50.7}   & 56.0  & \textbf{31.5} & 38.9   & 53.7  & 37.4\\
\bottomrule
\end{tabular}}
\caption{EE results on ACE05-E and CASIE. Bold numbers represent the highest score except for SFT approaches.  ($^*$ denotes selective instances)}
  \vspace{-0.5cm}
\label{tab:fs_ace}
\end{table*}

\paragraph{Baselines} 
{We consider the following  baselines that utilize zero-shot or in-context learning capabilities of LLMs:
(1) \textbf{ChatGPT-14}~\citep{li2023evaluating}, the first work that systematically analyzes the ChatGPT’s performance on information extraction (IE) tasks utilizing its zero-shot capabilities.
(2) \textbf{ChatGPT-IE}~\cite{han2023information}, which highlights that ChatGPT often generates longer trigger or argument spans, contributing to the evaluation gap between ChatGPT and tuning-based approaches. A soft-matching strategy is proposed to mitigate this evaluation gap, thereby providing a more accurate reflection of ChatGPT’s performance.
(3) \textbf{ChatIE}~\cite{wei2024chatie}, a multi-turn question-answering framework for zero-shot IE, wherein the first stage collects all the possible event types and in the second stage it performs information extraction for each event type. 
(4) \textbf{G-PTLM}~\cite{lin-etal-2023-global} regularize the event argument predictions by explicitly expressing argument constraints with prompts.
(5) \textbf{CODE4STRUCT}~\cite{wang-etal-2023-code4struct} formulate event extraction as a code generation problem, and represents event ontology in Python code expression.
(6) \textbf{Code4UIE}~\cite{guo2023retrieval}, another code generation-based approach, utilizing additional $M$ annotations retrieved from the training corpus with the highest similarity to the input sentence. The retrieved examples are used as ICL examples. 
}
In addition to the {zero-shot or in-context learning based approaches}, we include three supervised fine-tuning (SFT)
based approaches with relatively smaller LMs as baselines, including \textbf{DEGREE}~\cite{naacl2022degree}, \textbf{InstructUIE}~\cite{wang2023instructuie}, and \textbf{RexUIE}~\cite{liu-etal-2023-rexuie}.

\paragraph{Implementation Details} 
The proposed system is flexible, allowing any LLM to serve in any arbitrary agent role defined within the framework. In our experiments, we employ three LLMs: \texttt{Llama-3-8B-Instruct} (Llama3), \texttt{Gemini-Pro} (Gemini), and \texttt{GPT-3.5-turbo} (GPT). The results are presented under two distinct settings:
(a) Gemini-GPT: In this setting, two debaters are powered by Gemini and GPT, respectively. The Critic agent is powered by Gemini, while the Judge agent is powered by GPT.
(b) Llama3-GPT: Here, one debater uses \texttt{Llama-3-8B-Instruct} (Llama3), and the other uses \texttt{GPT-3.5-turbo} (GPT). Both the Critic and Judge agents are powered by Gemini.
We set the temperature of all agents to 0 to ensure reproducibility. Additional implementation details can be found in Appendix \ref{sec:exp_details}

\section{Results and Discussion}

\subsection{Main results}
The main results for ACE05 and CASIE are summarized in Table \ref{tab:fs_ace}. 
Aligned with previous observations, the performance gap {persists} between the proposed framework and advanced tuning-based methods. However, we emphasize that the gap is much smaller. 
For example on CASIE, the gap on ED shrinks by 17.9\% of the SOTA SFT baseline, and the system gains absolute 19.9\% F1 score gain on EAE over the Code4UIE baseline.
The performance gain over Code4UIE comes from three key aspects: the multi-agent debate system that leverages active discussion among agents, the effective utilization of ontology information, and the improved selection of relevant sentences. The detailed contribution of each component will be discussed in Section \ref{sec:ablation}.
{Regarding ontology usage, previous experimental results demonstrate consistent performance gains when ontology information is utilized. Our experimental results indicate that integrating the entire ontology schema information into the prompts cannot guarantee an optimal comprehension of the event schema by LLMs. Additionally, retrieving event information only for the types mentioned by the debaters is more computationally efficient.}

{Comparing the two different settings of LLM engines, Gemini-GPT and Llama3-GPT, their performance on ACE05 is relatively close. However, Llama3-GPT shows less promising performance on CASIE. This discrepancy arises because both GPT and Llama3 tend to generate longer spans. In ACE05, triggers are predefined to be one token, allowing GPT and Llama3 to follow instructions without generating long spans for event triggers. However, for arguments in ACE05 and both triggers and arguments in CASIE, GPT and Llama3 generate longer spans. For example, in CASIE, the average span length for Gemini is 9.0 tokens, while it is 13.7 tokens for GPT and 13.0 tokens for Llama3. Given that the average ground truth length of argument spans is 10.4 tokens, the argument spans generated by GPT and Llama3 are excessively long.}

\begin{figure*}[t!]
    \centering
    \begin{subfigure}[t]{0.32\textwidth}
        \centering
        \includegraphics[height=1.4in]{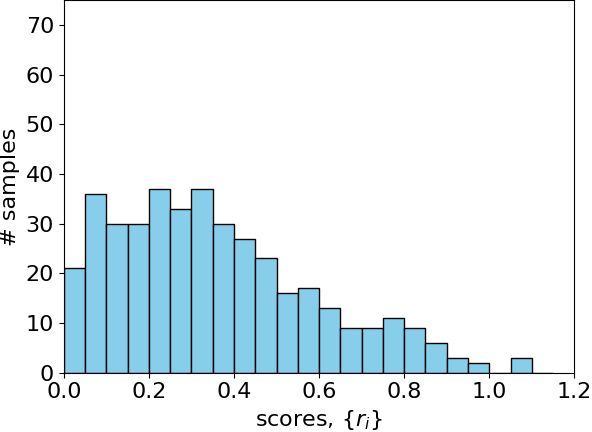}
        \caption{Before debate (ICL)}
         \vspace{-0.2cm}
    \end{subfigure}%
    \hfill
    \begin{subfigure}[t]{0.32\textwidth}
        \centering
        \includegraphics[height=1.4in]{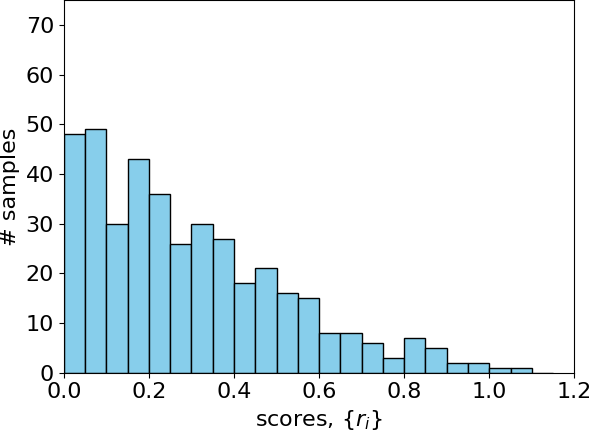}
        \caption{After 1st round debate}
         \vspace{-0.2cm}
    \end{subfigure}
    \hfill
    \begin{subfigure}[t]{0.32\textwidth}
        \centering
        \includegraphics[height=1.4in]{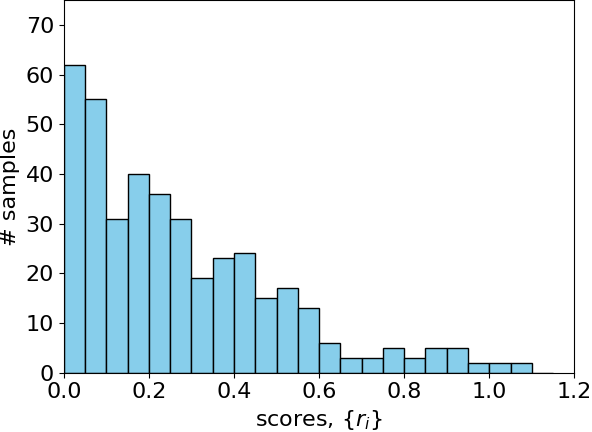}
        \caption{After 2nd round debate}
         \vspace{-0.2cm}
    \end{subfigure}
    \caption{Risk distribution evolution over the debate process}
    \label{fig:risk}
\end{figure*}

{Furthermore, we illustrate the evolution of the generation risk distribution throughout the debating process in Figure \ref{fig:risk}. The risk is measured by the calibration model, indicating the confidence (expressed by negative likelihood) of the LM generating the accurate answer given the input sentence and retrieved information. Initially, the risk distribution shows less confidence in accurate answers, as only ICL examples are available. As the debate progresses and more examples are retrieved, the model becomes more confident, which aligns with the findings in \cite{kang2024crag}. The risk distribution evolution visualizes the optimization of the event extraction outputs with the proposed retrieval module and validates the efficacy of the risk threshold decay strategy.}

\subsection{Ablation Study}
\label{sec:ablation}
To evaluate the effectiveness of each proposed module, an ablation study is conducted on ACE05 for 4 scenarios: without re-clustering, without the entire DRAG retrieval module, without AdaCP, and without both DRAG and AdaCP. The results are summarized in Table \ref{tab:ablation}.

\begin{table}[t!]
\centering
\scalebox{0.7}{
\begin{tabular}{l|c c|c c}
\toprule
\multirow{1}{*}{\textbf{Method}}&Ontology & Paradigm & ED & EAE\\
\midrule
\textbf{ChatGPT-IE}
            & \nomark   & ICL-5
            & 27.3   & 31.6 \\
\textbf{Code4UIE}  
            & \yesmark  & ICL-10$^*$
            & 37.4      & 57.0\\
\midrule
    \model          &\yesmark   &ICL-10$^*$ &50.2   & 59.5 \\
    - re-clustering &\yesmark   &ICL-10$^*$ &45.1   & 55.0 \\
    - DRAG          &\yesmark   &ICL-5  &39.9   & 52.8 \\
    - Calib         & \yesmark  &ICL-10$^*$  &40.6   & 57.3 \\
    - DRAG, Calib   & \yesmark  &ICL-5  &36.8   & 49.4 \\
\bottomrule
\end{tabular}}
\caption{Abalation study results}
  \vspace{-0.5cm}
\label{tab:ablation}
\end{table}

From the ablation study results, we may conclude that the integration of both the DRAG and AdaCP modules into a debating system significantly enhances event extraction performance.
Without the DRAG and AdaCP modules, the framework regresses to a basic debating system. However, this basic system still outperforms baseline approaches. 
This superiority arises from the ability of the debating system to capitalize on cross-examination capabilities among agents. Especially, the Critic agent gains the most effect during the cross-examination process. From 40 randomly sampled inferences from ACE05, the Critic improves 15\% of the event trigger answers. 

{In the absence of the DRAG module, the system regress to retrieving the closest data entries in the semantic space as reference data. 
The observed substantial performance degradation emphasizes the critical importance of incorporating diverse references for event extraction. 
Example (a) in Table \ref{tab:case} demonstrates how the DRAG module effectively corrects the event trigger token from ``holding'' to ``formerly''. Initially, the debater correctly identifies the event type as \texttt{Personal:Start-Position}, but mistakenly selects the verb ``holding'' as the event trigger. This is a common error in the first round of debate since early retrievals tend to favor verbs. Given the identified event type, more fine-grained reference data are retrieved, as shown in example (a),  which helps correctly identify ``formerly'' as the trigger. This underscores the effectiveness of the precise retrieval powered by the DRAG module.}

{Additionally, both ED and EAE show performance regression without the AdaCP module, especially for ED. Example (b) in Table \ref{tab:case} illustrates a case where the AdaCP module successfully rejects an incorrect ED result.  
Although the token "split" can imply a \texttt{Life:Divorce} event, the retrieved event definition "officially divorced under the legal definition of divorce" impacts the calibration model's confidence in its detection, successfully disambiguating it from a valid event mention. This example underscores the importance of the AdaCP in maintaining high detection accuracy.}

\begin{table*}[ht!]
\begin{center}\scalebox{0.75}{
\begin{tabular}{cp{6.5cm}p{10cm}p{2.5cm}}
\toprule
\textbf{ID} & Text & Conversations & GTH\\
\midrule
(a) 
& 
McCarthy was formerly a top civil servant at the Department of Trade and Industry. & \textbf{Debater:} ["Personnel:Start-Position", "holding"] \textbf{Retrieval:} 
- Example: "... and his successor as house majority whip and his former deputy ..." Answer: ["Personnel:End-Position", "former"]
&  ["Personnel:End-Position", "formerly"]\\\midrule
(b) & The celebrity couple spit up very publicly four years ago and each has since had well-publicized relationships with others .
&  \textbf{Debater:} ["Life:Divorce", "split"] \textbf{DRAG}: Life:Divorce: officially divorced under the legal definition of divorce \textbf{AdaCalib} (Answer fails calibration) -> [] & []\\

\bottomrule
\end{tabular}}

\caption{Examples illustrating the effect of DRAG and AdaCalib (Conversations are truncated for illustration).}
\vspace{-0.5cm}
\label{tab:error}
\end{center}
\end{table*}

\begin{table*}[ht!]
\begin{center}
{
\scalebox{0.7}{
\begin{tabular}{cp{9.5cm}p{4cm}p{7cm}}
\toprule
ID& Text & GTH & Predictions \\
\midrule
(a)&" We are studying that plan, we are examining it with our friends and allies, " Powell said, adding that \textbf{talks [Contact:Meet]} were now underway with the South Korean, Japanese, Russian and Australian as well as other governments. & Entity: governments 
& Entity: South Korean, Japanese, Russian, Australian, governments\\\midrule
(b)&The premier of the western Canadian province of British Columbia pleaded no contest to driving drunk during a Hawaiian \textbf{vacation [Movement:Transport]} in January. & Destination: Hawaiian & Destination: Hawaii \\\midrule
(c)&Does the threat posed by the Iraqi dictator justify a \textbf{war [Life:Attack]}, which is sure to \textbf{kill[Life:Die]} thousands of innocent children, women and men ? &
[Life:Die] Victim: men, Victim: women, Victim: children
&
[Life:Attack] Target: innocent children, women and men;
[Life:Die] Victim: thousands of innocent children, women and men\\ 

\bottomrule
\end{tabular}
}}  
\caption{Evaluation gap for LLMs (a-b) and challenging examples (c).}
\vspace{-0.2cm}
\label{tab:case}
\end{center}
\end{table*}
\subsection{Case Study}

The imperative for comprehensive argument extraction evaluation is underscored by our observations. While LLMs tend to identify longer spans than annotated arguments, this phenomenon does not necessarily reflect increased human-likeness in responses \cite{han2023information}. Rather, it often stems from underlying confusion regarding argument role spans.
Most prior supervised methods rely on evaluating exact matches of the head token of argument spans, owing to the challenges associated with assessing the entire argument extent. However, such an approach can yield inferior evaluations. Consider example (a) in Table \ref{tab:error}, where the argument extent of an \texttt{Entity} involved in the \texttt{Contact:Meet} event encompasses ``the South Korean, Japanese, Russian, and Australian as well as other governments'', with the head token being ``governments''. Existing evaluations based solely on the head token may overlook the nuanced understanding captured by the framework, which correctly predicts all governments attending the talks.
Thus, we advocate considering the entire argument's extent for precise evaluation, especially in the era of LLMs.

Token-level over-inference poses a challenge to the accuracy of current evaluation systems, particularly in reflecting the correctness of answers inferred from contextual clues. Consider example (b), where the correct argument role should encompass a word span from the original context. In this instance, the annotated argument role is ``Hawaiian'', while the predicted answer is ``Hawaii''. Although the answer is derived from the word ``Hawaiian'', it does not correspond to a valid token from the original sentence.
This observation underscores the necessity for more reference annotations in the event extraction task. By providing richer contextual cues, additional reference annotations can help mitigate token-level over-inference and enhance the precision of evaluations.

In the context of example (c), the framework demonstrates accurate prediction of the victims of the \texttt{Life:Die} event (regardless of the span confusion mentioned in (a)), encompassing ``men'', ``women'', and ``children''. However, it overpredicts the target of the war as ``innocent children, women, and men''. Despite encountering numerous examples with closely aligned semantic meanings, including instances where the trigger token is also ``war'', the system struggles to differentiate between the target for the ``war'' event and individuals affected by the ``war''. It highlights that the current guidelines and contextual examples remain insufficient to fully address the reasoning behind such occurrences. 
\section{Conclusion}

This work introduces a novel multi-agent debate paradigm that resembles the optimization process. This debate model is conceptualized as an optimization mechanism wherein supporting information is systematically retrieved to regulate the distribution of risk. The evolution of risk distribution throughout the debating process illustrates how the integration of the adaptive conformal prediction module and the diverse RAG module can progressively steer the risk distribution towards more confident answers. Through this framework, the debate process becomes not just a discourse but a strategic endeavor aimed at achieving optimal outcomes.
\section*{Limitations}

In this work, we found that leveraging multi-agent debating to iteratively refine the event extraction output without tuning LLMs leads to significant performance gains for LLM-based in-context learning (ICL) on event extraction. We are particularly excited about the system's ability to effortlessly adapt to new domains or ontologies. However, compared to previous zero-shot or ICL event extraction approaches, our proposed system requires multiple rounds of LLM inferences, increasing both inference time and cost. We welcome follow-up work and optimization, as we believe many of these issues can be addressed.

\bibliography{ref}

\begin{thebibliography}{61}
\providecommand{\natexlab}[1]{#1}

\bibitem[{Ahn(2006)}]{ahn2006stages}
David Ahn. 2006.
\newblock The stages of event extraction.
\newblock In \emph{Proceedings of the Workshop on Annotating and Reasoning about Time and Events}, pages 1--8.

\bibitem[{AI@Meta(2024)}]{llama3modelcard}
AI@Meta. 2024.
\newblock \href {https://github.com/meta-llama/llama3/blob/main/MODEL_CARD.md} {Llama 3 model card}.

\bibitem[{Angelopoulos et~al.(2022)Angelopoulos, Bates, Candès, Jordan, and Lei}]{angelopoulos2022learn}
Anastasios~N. Angelopoulos, Stephen Bates, Emmanuel~J. Candès, Michael~I. Jordan, and Lihua Lei. 2022.
\newblock \href {https://arxiv.org/abs/2110.01052} {Learn then test: Calibrating predictive algorithms to achieve risk control}.
\newblock \emph{Preprint}, arXiv:2110.01052.

\bibitem[{Anil et~al.(2023)Anil, Dai, Firat, Johnson, Lepikhin, Passos, Shakeri, Taropa, Bailey, Chen, Chu, Clark, Shafey, Huang, Meier-Hellstern, Mishra, Moreira, Omernick, Robinson, Ruder, Tay, Xiao, Xu, Zhang, Abrego, Ahn, Austin, Barham, Botha, Bradbury, Brahma, Brooks, Catasta, Cheng, Cherry, Choquette-Choo, Chowdhery, Crepy, Dave, Dehghani, Dev, Devlin, Díaz, Du, Dyer, Feinberg, Feng, Fienber, Freitag, Garcia, Gehrmann, Gonzalez, Gur-Ari, Hand, Hashemi, Hou, Howland, Hu, Hui, Hurwitz, Isard, Ittycheriah, Jagielski, Jia, Kenealy, Krikun, Kudugunta, Lan, Lee, Lee, Li, Li, Li, Li, Li, Lim, Lin, Liu, Liu, Maggioni, Mahendru, Maynez, Misra, Moussalem, Nado, Nham, Ni, Nystrom, Parrish, Pellat, Polacek, Polozov, Pope, Qiao, Reif, Richter, Riley, Ros, Roy, Saeta, Samuel, Shelby, Slone, Smilkov, So, Sohn, Tokumine, Valter, Vasudevan, Vodrahalli, Wang, Wang, Wang, Wang, Wieting, Wu, Xu, Xu, Xue, Yin, Yu, Zhang, Zheng, Zheng, Zhou, Zhou, Petrov, and Wu}]{anil2023palm}
Rohan Anil, Andrew~M. Dai, Orhan Firat, Melvin Johnson, Dmitry Lepikhin, Alexandre Passos, Siamak Shakeri, Emanuel Taropa, Paige Bailey, Zhifeng Chen, Eric Chu, Jonathan~H. Clark, Laurent~El Shafey, Yanping Huang, Kathy Meier-Hellstern, Gaurav Mishra, Erica Moreira, Mark Omernick, Kevin Robinson, Sebastian Ruder, Yi~Tay, Kefan Xiao, Yuanzhong Xu, Yujing Zhang, Gustavo~Hernandez Abrego, Junwhan Ahn, Jacob Austin, Paul Barham, Jan Botha, James Bradbury, Siddhartha Brahma, Kevin Brooks, Michele Catasta, Yong Cheng, Colin Cherry, Christopher~A. Choquette-Choo, Aakanksha Chowdhery, Clément Crepy, Shachi Dave, Mostafa Dehghani, Sunipa Dev, Jacob Devlin, Mark Díaz, Nan Du, Ethan Dyer, Vlad Feinberg, Fangxiaoyu Feng, Vlad Fienber, Markus Freitag, Xavier Garcia, Sebastian Gehrmann, Lucas Gonzalez, Guy Gur-Ari, Steven Hand, Hadi Hashemi, Le~Hou, Joshua Howland, Andrea Hu, Jeffrey Hui, Jeremy Hurwitz, Michael Isard, Abe Ittycheriah, Matthew Jagielski, Wenhao Jia, Kathleen Kenealy, Maxim Krikun, Sneha Kudugunta, Chang
  Lan, Katherine Lee, Benjamin Lee, Eric Li, Music Li, Wei Li, YaGuang Li, Jian Li, Hyeontaek Lim, Hanzhao Lin, Zhongtao Liu, Frederick Liu, Marcello Maggioni, Aroma Mahendru, Joshua Maynez, Vedant Misra, Maysam Moussalem, Zachary Nado, John Nham, Eric Ni, Andrew Nystrom, Alicia Parrish, Marie Pellat, Martin Polacek, Alex Polozov, Reiner Pope, Siyuan Qiao, Emily Reif, Bryan Richter, Parker Riley, Alex~Castro Ros, Aurko Roy, Brennan Saeta, Rajkumar Samuel, Renee Shelby, Ambrose Slone, Daniel Smilkov, David~R. So, Daniel Sohn, Simon Tokumine, Dasha Valter, Vijay Vasudevan, Kiran Vodrahalli, Xuezhi Wang, Pidong Wang, Zirui Wang, Tao Wang, John Wieting, Yuhuai Wu, Kelvin Xu, Yunhan Xu, Linting Xue, Pengcheng Yin, Jiahui Yu, Qiao Zhang, Steven Zheng, Ce~Zheng, Weikang Zhou, Denny Zhou, Slav Petrov, and Yonghui Wu. 2023.
\newblock \href {https://arxiv.org/abs/2305.10403} {Palm 2 technical report}.
\newblock \emph{Preprint}, arXiv:2305.10403.

\bibitem[{Asai et~al.(2024)Asai, Wu, Wang, Sil, and Hajishirzi}]{asai2024selfrag}
Akari Asai, Zeqiu Wu, Yizhong Wang, Avirup Sil, and Hannaneh Hajishirzi. 2024.
\newblock \href {https://openreview.net/forum?id=hSyW5go0v8} {Self-{RAG}: Learning to retrieve, generate, and critique through self-reflection}.
\newblock In \emph{The Twelfth International Conference on Learning Representations}.

\bibitem[{Bates et~al.(2021)Bates, Angelopoulos, Lei, Malik, and Jordan}]{10.1145/3478535}
Stephen Bates, Anastasios Angelopoulos, Lihua Lei, Jitendra Malik, and Michael Jordan. 2021.
\newblock \href {https://doi.org/10.1145/3478535} {Distribution-free, risk-controlling prediction sets}.
\newblock \emph{J. ACM}, 68(6).

\bibitem[{Brown et~al.(2020)Brown, Mann, Ryder, Subbiah, Kaplan, Dhariwal, Neelakantan, Shyam, Sastry, Askell, Agarwal, Herbert-Voss, Krueger, Henighan, Child, Ramesh, Ziegler, Wu, Winter, Hesse, Chen, Sigler, Litwin, Gray, Chess, Clark, Berner, McCandlish, Radford, Sutskever, and Amodei}]{brown2020language}
Tom~B. Brown, Benjamin Mann, Nick Ryder, Melanie Subbiah, Jared Kaplan, Prafulla Dhariwal, Arvind Neelakantan, Pranav Shyam, Girish Sastry, Amanda Askell, Sandhini Agarwal, Ariel Herbert-Voss, Gretchen Krueger, Tom Henighan, Rewon Child, Aditya Ramesh, Daniel~M. Ziegler, Jeffrey Wu, Clemens Winter, Christopher Hesse, Mark Chen, Eric Sigler, Mateusz Litwin, Scott Gray, Benjamin Chess, Jack Clark, Christopher Berner, Sam McCandlish, Alec Radford, Ilya Sutskever, and Dario Amodei. 2020.
\newblock \href {https://arxiv.org/abs/2005.14165} {Language models are few-shot learners}.
\newblock \emph{Preprint}, arXiv:2005.14165.

\bibitem[{Chan et~al.(2023)Chan, Chen, Su, Yu, Xue, Zhang, Fu, and Liu}]{chan2023chateval}
Chi-Min Chan, Weize Chen, Yusheng Su, Jianxuan Yu, Wei Xue, Shanghang Zhang, Jie Fu, and Zhiyuan Liu. 2023.
\newblock \href {https://arxiv.org/abs/2308.07201} {Chateval: Towards better llm-based evaluators through multi-agent debate}.
\newblock \emph{Preprint}, arXiv:2308.07201.

\bibitem[{Chen et~al.(2024)Chen, Lin, Han, and Sun}]{0011LH024}
Jiawei Chen, Hongyu Lin, Xianpei Han, and Le~Sun. 2024.
\newblock \href {https://doi.org/10.1609/AAAI.V38I16.29728} {Benchmarking large language models in retrieval-augmented generation}.
\newblock In \emph{Thirty-Eighth {AAAI} Conference on Artificial Intelligence, {AAAI} 2024, Thirty-Sixth Conference on Innovative Applications of Artificial Intelligence, {IAAI} 2024, Fourteenth Symposium on Educational Advances in Artificial Intelligence, {EAAI} 2014, February 20-27, 2024, Vancouver, Canada}, pages 17754--17762. {AAAI} Press.

\bibitem[{Chen et~al.(2023)Chen, Su, Zuo, Yang, Yuan, Chan, Yu, Lu, Hung, Qian, Qin, Cong, Xie, Liu, Sun, and Zhou}]{chen2023agentverse}
Weize Chen, Yusheng Su, Jingwei Zuo, Cheng Yang, Chenfei Yuan, Chi-Min Chan, Heyang Yu, Yaxi Lu, Yi-Hsin Hung, Chen Qian, Yujia Qin, Xin Cong, Ruobing Xie, Zhiyuan Liu, Maosong Sun, and Jie Zhou. 2023.
\newblock \href {https://arxiv.org/abs/2308.10848} {Agentverse: Facilitating multi-agent collaboration and exploring emergent behaviors}.
\newblock \emph{Preprint}, arXiv:2308.10848.

\bibitem[{Chen et~al.(2022)Chen, Hu, Chen, Verga, and Cohen}]{chen-etal-2022-murag}
Wenhu Chen, Hexiang Hu, Xi~Chen, Pat Verga, and William Cohen. 2022.
\newblock \href {https://doi.org/10.18653/v1/2022.emnlp-main.375} {{M}u{RAG}: Multimodal retrieval-augmented generator for open question answering over images and text}.
\newblock In \emph{Proceedings of the 2022 Conference on Empirical Methods in Natural Language Processing}, pages 5558--5570, Abu Dhabi, United Arab Emirates. Association for Computational Linguistics.

\bibitem[{Chinchor and Marsh(1998)}]{chinchor1998muc}
Nancy Chinchor and Elaine Marsh. 1998.
\newblock Muc-7 information extraction task definition.
\newblock In \emph{Proceeding of the seventh message understanding conference (MUC-7), Appendices}, pages 359--367.

\bibitem[{Chowdhery et~al.(2022)Chowdhery, Narang, Devlin, Bosma, Mishra, Roberts, Barham, Chung, Sutton, Gehrmann, Schuh, Shi, Tsvyashchenko, Maynez, Rao, Barnes, Tay, Shazeer, Prabhakaran, Reif, Du, Hutchinson, Pope, Bradbury, Austin, Isard, Gur-Ari, Yin, Duke, Levskaya, Ghemawat, Dev, Michalewski, Garcia, Misra, Robinson, Fedus, Zhou, Ippolito, Luan, Lim, Zoph, Spiridonov, Sepassi, Dohan, Agrawal, Omernick, Dai, Pillai, Pellat, Lewkowycz, Moreira, Child, Polozov, Lee, Zhou, Wang, Saeta, Diaz, Firat, Catasta, Wei, Meier-Hellstern, Eck, Dean, Petrov, and Fiedel}]{chowdhery2022palm}
Aakanksha Chowdhery, Sharan Narang, Jacob Devlin, Maarten Bosma, Gaurav Mishra, Adam Roberts, Paul Barham, Hyung~Won Chung, Charles Sutton, Sebastian Gehrmann, Parker Schuh, Kensen Shi, Sasha Tsvyashchenko, Joshua Maynez, Abhishek Rao, Parker Barnes, Yi~Tay, Noam Shazeer, Vinodkumar Prabhakaran, Emily Reif, Nan Du, Ben Hutchinson, Reiner Pope, James Bradbury, Jacob Austin, Michael Isard, Guy Gur-Ari, Pengcheng Yin, Toju Duke, Anselm Levskaya, Sanjay Ghemawat, Sunipa Dev, Henryk Michalewski, Xavier Garcia, Vedant Misra, Kevin Robinson, Liam Fedus, Denny Zhou, Daphne Ippolito, David Luan, Hyeontaek Lim, Barret Zoph, Alexander Spiridonov, Ryan Sepassi, David Dohan, Shivani Agrawal, Mark Omernick, Andrew~M. Dai, Thanumalayan~Sankaranarayana Pillai, Marie Pellat, Aitor Lewkowycz, Erica Moreira, Rewon Child, Oleksandr Polozov, Katherine Lee, Zongwei Zhou, Xuezhi Wang, Brennan Saeta, Mark Diaz, Orhan Firat, Michele Catasta, Jason Wei, Kathy Meier-Hellstern, Douglas Eck, Jeff Dean, Slav Petrov, and Noah Fiedel. 2022.
\newblock \href {https://arxiv.org/abs/2204.02311} {Palm: Scaling language modeling with pathways}.
\newblock \emph{Preprint}, arXiv:2204.02311.

\bibitem[{Du and Cardie(2020)}]{xinyaduEMNLP2020}
Xinya Du and Claire Cardie. 2020.
\newblock \href {https://doi.org/10.18653/v1/2020.emnlp-main.49} {Event extraction by answering (almost) natural questions}.
\newblock In \emph{Proceedings of the 2020 Conference on Empirical Methods in Natural Language Processing (EMNLP)}, pages 671--683, Online. Association for Computational Linguistics.

\bibitem[{Du et~al.(2023)Du, Li, Torralba, Tenenbaum, and Mordatch}]{du2023improving}
Yilun Du, Shuang Li, Antonio Torralba, Joshua~B. Tenenbaum, and Igor Mordatch. 2023.
\newblock \href {https://arxiv.org/abs/2305.14325} {Improving factuality and reasoning in language models through multiagent debate}.
\newblock \emph{Preprint}, arXiv:2305.14325.

\bibitem[{Fu et~al.(2023)Fu, Peng, Khot, and Lapata}]{fu2023improving}
Yao Fu, Hao Peng, Tushar Khot, and Mirella Lapata. 2023.
\newblock \href {https://arxiv.org/abs/2305.10142} {Improving language model negotiation with self-play and in-context learning from ai feedback}.
\newblock \emph{Preprint}, arXiv:2305.10142.

\bibitem[{Gammerman et~al.(1998)Gammerman, Vovk, and Vapnik}]{10.5555/2074094.2074112}
A.~Gammerman, V.~Vovk, and V.~Vapnik. 1998.
\newblock Learning by transduction.
\newblock In \emph{Proceedings of the Fourteenth Conference on Uncertainty in Artificial Intelligence}, UAI'98, page 148–155, San Francisco, CA, USA. Morgan Kaufmann Publishers Inc.

\bibitem[{Gao et~al.(2023)Gao, Zhao, Yu, and Xu}]{gao2023exploringChatGPTee}
Jun Gao, Huan Zhao, Changlong Yu, and Ruifeng Xu. 2023.
\newblock \href {https://arxiv.org/abs/2303.03836} {Exploring the feasibility of chatgpt for event extraction}.
\newblock \emph{Preprint}, arXiv:2303.03836.

\bibitem[{Glass et~al.(2022)Glass, Rossiello, Chowdhury, Naik, Cai, and Gliozzo}]{glass-etal-2022-re2g}
Michael Glass, Gaetano Rossiello, Md~Faisal~Mahbub Chowdhury, Ankita Naik, Pengshan Cai, and Alfio Gliozzo. 2022.
\newblock \href {https://doi.org/10.18653/v1/2022.naacl-main.194} {{R}e2{G}: Retrieve, rerank, generate}.
\newblock In \emph{Proceedings of the 2022 Conference of the North American Chapter of the Association for Computational Linguistics: Human Language Technologies}, pages 2701--2715, Seattle, United States. Association for Computational Linguistics.

\bibitem[{Grishman(1997)}]{grishman1997information}
Ralph Grishman. 1997.
\newblock Information extraction: Techniques and challenges.
\newblock In \emph{International summer school on information extraction}, pages 10--27. Springer.

\bibitem[{Guo et~al.(2023)Guo, Li, Jin, Liu, Zeng, Liu, Li, Yang, Bai, Guo et~al.}]{guo2023retrieval}
Yucan Guo, Zixuan Li, Xiaolong Jin, Yantao Liu, Yutao Zeng, Wenxuan Liu, Xiang Li, Pan Yang, Long Bai, Jiafeng Guo, et~al. 2023.
\newblock Retrieval-augmented code generation for universal information extraction.
\newblock \emph{arXiv preprint arXiv:2311.02962}.

\bibitem[{Han et~al.(2023)Han, Peng, Yang, Wang, Liu, and Wan}]{han2023information}
Ridong Han, Tao Peng, Chaohao Yang, Benyou Wang, Lu~Liu, and Xiang Wan. 2023.
\newblock \href {https://arxiv.org/abs/2305.14450} {Is information extraction solved by chatgpt? an analysis of performance, evaluation criteria, robustness and errors}.
\newblock \emph{Preprint}, arXiv:2305.14450.

\bibitem[{Hsu et~al.(2022{\natexlab{a}})Hsu, Huang, Boschee, Miller, Natarajan, Chang, and Peng}]{hsu-etal-2022-degree}
I-Hung Hsu, Kuan-Hao Huang, Elizabeth Boschee, Scott Miller, Prem Natarajan, Kai-Wei Chang, and Nanyun Peng. 2022{\natexlab{a}}.
\newblock \href {https://doi.org/10.18653/v1/2022.naacl-main.138} {{DEGREE}: A data-efficient generation-based event extraction model}.
\newblock In \emph{Proceedings of the 2022 Conference of the North American Chapter of the Association for Computational Linguistics: Human Language Technologies}, pages 1890--1908, Seattle, United States. Association for Computational Linguistics.

\bibitem[{Hsu et~al.(2022{\natexlab{b}})Hsu, Huang, Boschee, Miller, Natarajan, Chang, and Peng}]{naacl2022degree}
I-Hung Hsu, Kuan-Hao Huang, Elizabeth Boschee, Scott Miller, Prem Natarajan, Kai-Wei Chang, and Nanyun Peng. 2022{\natexlab{b}}.
\newblock Degree: A data-efficient generative event extraction model.
\newblock In \emph{Proceedings of the 2022 Conference of the North American Chapter of the Association for Computational Linguistics (NAACL)}.

\bibitem[{Jiang et~al.(2023)Jiang, Xu, Gao, Sun, Liu, Dwivedi-Yu, Yang, Callan, and Neubig}]{jiang-etal-2023-active}
Zhengbao Jiang, Frank Xu, Luyu Gao, Zhiqing Sun, Qian Liu, Jane Dwivedi-Yu, Yiming Yang, Jamie Callan, and Graham Neubig. 2023.
\newblock \href {https://doi.org/10.18653/v1/2023.emnlp-main.495} {Active retrieval augmented generation}.
\newblock In \emph{Proceedings of the 2023 Conference on Empirical Methods in Natural Language Processing}, pages 7969--7992, Singapore. Association for Computational Linguistics.

\bibitem[{Jing~Lei and Wasserman(2013)}]{10.1080/01621459.2012.751873}
James~Robins Jing~Lei and Larry Wasserman. 2013.
\newblock \href {https://doi.org/10.1080/01621459.2012.751873} {Distribution-free prediction sets}.
\newblock \emph{Journal of the American Statistical Association}, 108(501):278--287.
\newblock PMID: 25237208.

\bibitem[{Kang et~al.(2024)Kang, Gürel, Yu, Song, and Li}]{kang2024crag}
Mintong Kang, Nezihe~Merve Gürel, Ning Yu, Dawn Song, and Bo~Li. 2024.
\newblock \href {https://arxiv.org/abs/2402.03181} {C-rag: Certified generation risks for retrieval-augmented language models}.
\newblock \emph{Preprint}, arXiv:2402.03181.

\bibitem[{Lewis et~al.(2020)Lewis, Perez, Piktus, Petroni, Karpukhin, Goyal, K\"{u}ttler, Lewis, Yih, Rockt\"{a}schel, Riedel, and Kiela}]{10.5555/3495724.3496517}
Patrick Lewis, Ethan Perez, Aleksandra Piktus, Fabio Petroni, Vladimir Karpukhin, Naman Goyal, Heinrich K\"{u}ttler, Mike Lewis, Wen-tau Yih, Tim Rockt\"{a}schel, Sebastian Riedel, and Douwe Kiela. 2020.
\newblock Retrieval-augmented generation for knowledge-intensive nlp tasks.
\newblock In \emph{Proceedings of the 34th International Conference on Neural Information Processing Systems}, NIPS '20, Red Hook, NY, USA. Curran Associates Inc.

\bibitem[{Li et~al.(2023{\natexlab{a}})Li, Fang, Yang, Wang, Ye, Zhao, and Zhang}]{li2023evaluating}
Bo~Li, Gexiang Fang, Yang Yang, Quansen Wang, Wei Ye, Wen Zhao, and Shikun Zhang. 2023{\natexlab{a}}.
\newblock \href {https://arxiv.org/abs/2304.11633} {Evaluating chatgpt's information extraction capabilities: An assessment of performance, explainability, calibration, and faithfulness}.
\newblock \emph{Preprint}, arXiv:2304.11633.

\bibitem[{Li et~al.(2023{\natexlab{b}})Li, Tang, Zhao, Wang, Nie, and Wen}]{li-etal-2023-web}
Junyi Li, Tianyi Tang, Wayne~Xin Zhao, Jingyuan Wang, Jian-Yun Nie, and Ji-Rong Wen. 2023{\natexlab{b}}.
\newblock \href {https://doi.org/10.18653/v1/2023.findings-acl.46} {The web can be your oyster for improving language models}.
\newblock In \emph{Findings of the Association for Computational Linguistics: ACL 2023}, pages 728--746, Toronto, Canada. Association for Computational Linguistics.

\bibitem[{Liang et~al.(2023)Liang, He, Jiao, Wang, Wang, Wang, Yang, Tu, and Shi}]{liang2023encouraging}
Tian Liang, Zhiwei He, Wenxiang Jiao, Xing Wang, Yan Wang, Rui Wang, Yujiu Yang, Zhaopeng Tu, and Shuming Shi. 2023.
\newblock \href {https://arxiv.org/abs/2305.19118} {Encouraging divergent thinking in large language models through multi-agent debate}.
\newblock \emph{Preprint}, arXiv:2305.19118.

\bibitem[{Lin et~al.(2020)Lin, Ji, Huang, and Wu}]{yinglinACL2020}
Ying Lin, Heng Ji, Fei Huang, and Lingfei Wu. 2020.
\newblock \href {https://doi.org/10.18653/v1/2020.acl-main.713} {A joint neural model for information extraction with global features}.
\newblock In \emph{Proceedings of the 58th Annual Meeting of the Association for Computational Linguistics}, pages 7999--8009, Online. Association for Computational Linguistics.

\bibitem[{Lin et~al.(2023)Lin, Zhang, and Song}]{lin-etal-2023-global}
Zizheng Lin, Hongming Zhang, and Yangqiu Song. 2023.
\newblock \href {https://doi.org/10.18653/v1/2023.findings-eacl.191} {Global constraints with prompting for zero-shot event argument classification}.
\newblock In \emph{Findings of the Association for Computational Linguistics: EACL 2023}, pages 2527--2538, Dubrovnik, Croatia. Association for Computational Linguistics.

\bibitem[{{Linguistic Data Consortium}(2005)}]{ldc_ace05}
{Linguistic Data Consortium}. 2005.
\newblock English annotation guidelines for events.
\newblock \url{https://www.ldc.upenn.edu/ sites/www.ldc.upenn.edu/files/ english-events-guidelines-v5.4.3. pdf.}

\bibitem[{Liu et~al.(2023)Liu, Zhao, Kang, Zhang, Zhou, Sun, Kuang, and Wu}]{liu-etal-2023-rexuie}
Chengyuan Liu, Fubang Zhao, Yangyang Kang, Jingyuan Zhang, Xiang Zhou, Changlong Sun, Kun Kuang, and Fei Wu. 2023.
\newblock \href {https://doi.org/10.18653/v1/2023.findings-emnlp.1024} {{R}ex{UIE}: A recursive method with explicit schema instructor for universal information extraction}.
\newblock In \emph{Findings of the Association for Computational Linguistics: EMNLP 2023}, pages 15342--15359, Singapore. Association for Computational Linguistics.

\bibitem[{OpenAI(2023{\natexlab{a}})}]{openai_chatgpt}
OpenAI. 2023{\natexlab{a}}.
\newblock Chatgpt: Openai's language model.
\newblock \url{https://openai.com/chatgpt}.
\newblock Accessed: November 10, 2023.

\bibitem[{OpenAI(2023{\natexlab{b}})}]{openai_gpt3}
OpenAI. 2023{\natexlab{b}}.
\newblock Gpt-3: Openai's language model.
\newblock Accessed: November 10, 2023.
\newblock Available at \url{https://www.openai.com/}.

\bibitem[{OpenAI(2023{\natexlab{c}})}]{openai_gpt4}
OpenAI. 2023{\natexlab{c}}.
\newblock Gpt-4 is openai’s most advanced system, producing safer and more useful responses.
\newblock Accessed: November 10, 2023.
\newblock Available at \url{https://openai.com/gpt-4}.

\bibitem[{Park et~al.(2023)Park, O'Brien, Cai, Morris, Liang, and Bernstein}]{park2023generative}
Joon~Sung Park, Joseph~C. O'Brien, Carrie~J. Cai, Meredith~Ringel Morris, Percy Liang, and Michael~S. Bernstein. 2023.
\newblock \href {https://arxiv.org/abs/2304.03442} {Generative agents: Interactive simulacra of human behavior}.
\newblock \emph{Preprint}, arXiv:2304.03442.

\bibitem[{Qian et~al.(2023)Qian, Cong, Liu, Yang, Chen, Su, Dang, Li, Xu, Li, Liu, and Sun}]{qian2023communicative}
Chen Qian, Xin Cong, Wei Liu, Cheng Yang, Weize Chen, Yusheng Su, Yufan Dang, Jiahao Li, Juyuan Xu, Dahai Li, Zhiyuan Liu, and Maosong Sun. 2023.
\newblock \href {https://arxiv.org/abs/2307.07924} {Communicative agents for software development}.
\newblock \emph{Preprint}, arXiv:2307.07924.

\bibitem[{Quach et~al.(2024)Quach, Fisch, Schuster, Yala, Sohn, Jaakkola, and Barzilay}]{quach2024conformal}
Victor Quach, Adam Fisch, Tal Schuster, Adam Yala, Jae~Ho Sohn, Tommi~S. Jaakkola, and Regina Barzilay. 2024.
\newblock \href {https://openreview.net/forum?id=pzUhfQ74c5} {Conformal language modeling}.
\newblock In \emph{The Twelfth International Conference on Learning Representations}.

\bibitem[{Satyapanich et~al.(2020)Satyapanich, Ferraro, and Finin}]{Satyapanich2020CASIEEC}
Taneeya Satyapanich, Francis Ferraro, and Timothy~W. Finin. 2020.
\newblock Casie: Extracting cybersecurity event information from text.
\newblock In \emph{AAAI Conference on Artificial Intelligence}.

\bibitem[{Shafer and Vovk(2008)}]{10.5555/1390681.1390693}
Glenn Shafer and Vladimir Vovk. 2008.
\newblock A tutorial on conformal prediction.
\newblock \emph{J. Mach. Learn. Res.}, 9:371–421.

\bibitem[{Siriwardhana et~al.(2023)Siriwardhana, Weerasekera, Wen, Kaluarachchi, Rana, and Nanayakkara}]{siriwardhana-etal-2023-improving}
Shamane Siriwardhana, Rivindu Weerasekera, Elliott Wen, Tharindu Kaluarachchi, Rajib Rana, and Suranga Nanayakkara. 2023.
\newblock \href {https://doi.org/10.1162/tacl_a_00530} {Improving the domain adaptation of retrieval augmented generation ({RAG}) models for open domain question answering}.
\newblock \emph{Transactions of the Association for Computational Linguistics}, 11:1--17.

\bibitem[{Song et~al.(2015)Song, Bies, Strassel, Riese, Mott, Ellis, Wright, Kulick, Ryant, and Ma}]{song2015light}
Zhiyi Song, Ann Bies, Stephanie Strassel, Tom Riese, Justin Mott, Joe Ellis, Jonathan Wright, Seth Kulick, Neville Ryant, and Xiaoyi Ma. 2015.
\newblock From light to rich ere: annotation of entities, relations, and events.
\newblock In \emph{Proceedings of the the 3rd Workshop on EVENTS: Definition, Detection, Coreference, and Representation}, pages 89--98.

\bibitem[{Touvron et~al.(2023{\natexlab{a}})Touvron, Lavril, Izacard, Martinet, Lachaux, Lacroix, Rozière, Goyal, Hambro, Azhar, Rodriguez, Joulin, Grave, and Lample}]{touvron2023llama}
Hugo Touvron, Thibaut Lavril, Gautier Izacard, Xavier Martinet, Marie-Anne Lachaux, Timothée Lacroix, Baptiste Rozière, Naman Goyal, Eric Hambro, Faisal Azhar, Aurelien Rodriguez, Armand Joulin, Edouard Grave, and Guillaume Lample. 2023{\natexlab{a}}.
\newblock \href {https://arxiv.org/abs/2302.13971} {Llama: Open and efficient foundation language models}.
\newblock \emph{Preprint}, arXiv:2302.13971.

\bibitem[{Touvron et~al.(2023{\natexlab{b}})Touvron, Martin, Stone, Albert, Almahairi, Babaei, Bashlykov, Batra, Bhargava, Bhosale, Bikel, Blecher, Ferrer, Chen, Cucurull, Esiobu, Fernandes, Fu, Fu, Fuller, Gao, Goswami, Goyal, Hartshorn, Hosseini, Hou, Inan, Kardas, Kerkez, Khabsa, Kloumann, Korenev, Koura, Lachaux, Lavril, Lee, Liskovich, Lu, Mao, Martinet, Mihaylov, Mishra, Molybog, Nie, Poulton, Reizenstein, Rungta, Saladi, Schelten, Silva, Smith, Subramanian, Tan, Tang, Taylor, Williams, Kuan, Xu, Yan, Zarov, Zhang, Fan, Kambadur, Narang, Rodriguez, Stojnic, Edunov, and Scialom}]{touvron2023llama2}
Hugo Touvron, Louis Martin, Kevin Stone, Peter Albert, Amjad Almahairi, Yasmine Babaei, Nikolay Bashlykov, Soumya Batra, Prajjwal Bhargava, Shruti Bhosale, Dan Bikel, Lukas Blecher, Cristian~Canton Ferrer, Moya Chen, Guillem Cucurull, David Esiobu, Jude Fernandes, Jeremy Fu, Wenyin Fu, Brian Fuller, Cynthia Gao, Vedanuj Goswami, Naman Goyal, Anthony Hartshorn, Saghar Hosseini, Rui Hou, Hakan Inan, Marcin Kardas, Viktor Kerkez, Madian Khabsa, Isabel Kloumann, Artem Korenev, Punit~Singh Koura, Marie-Anne Lachaux, Thibaut Lavril, Jenya Lee, Diana Liskovich, Yinghai Lu, Yuning Mao, Xavier Martinet, Todor Mihaylov, Pushkar Mishra, Igor Molybog, Yixin Nie, Andrew Poulton, Jeremy Reizenstein, Rashi Rungta, Kalyan Saladi, Alan Schelten, Ruan Silva, Eric~Michael Smith, Ranjan Subramanian, Xiaoqing~Ellen Tan, Binh Tang, Ross Taylor, Adina Williams, Jian~Xiang Kuan, Puxin Xu, Zheng Yan, Iliyan Zarov, Yuchen Zhang, Angela Fan, Melanie Kambadur, Sharan Narang, Aurelien Rodriguez, Robert Stojnic, Sergey Edunov, and Thomas
  Scialom. 2023{\natexlab{b}}.
\newblock \href {https://arxiv.org/abs/2307.09288} {Llama 2: Open foundation and fine-tuned chat models}.
\newblock \emph{Preprint}, arXiv:2307.09288.

\bibitem[{Vovk et~al.(2005)Vovk, Gammerman, and Shafer}]{10.5555/1062391}
Vladimir Vovk, Alex Gammerman, and Glenn Shafer. 2005.
\newblock \emph{Algorithmic Learning in a Random World}.
\newblock Springer-Verlag, Berlin, Heidelberg.

\bibitem[{Wadden et~al.(2019)Wadden, Wennberg, Luan, and Hajishirzi}]{wadden-etal-2019-entity}
David Wadden, Ulme Wennberg, Yi~Luan, and Hannaneh Hajishirzi. 2019.
\newblock \href {https://doi.org/10.18653/v1/D19-1585} {Entity, relation, and event extraction with contextualized span representations}.
\newblock In \emph{Proceedings of the 2019 Conference on Empirical Methods in Natural Language Processing and the 9th International Joint Conference on Natural Language Processing (EMNLP-IJCNLP)}, pages 5784--5789, Hong Kong, China. Association for Computational Linguistics.

\bibitem[{Wang et~al.(2023{\natexlab{a}})Wang, Du, Yu, Chen, Zhu, Chu, Yan, and Guan}]{wang2023apollos}
Haotian Wang, Xiyuan Du, Weijiang Yu, Qianglong Chen, Kun Zhu, Zheng Chu, Lian Yan, and Yi~Guan. 2023{\natexlab{a}}.
\newblock \href {https://arxiv.org/abs/2312.04854} {Apollo's oracle: Retrieval-augmented reasoning in multi-agent debates}.
\newblock \emph{Preprint}, arXiv:2312.04854.

\bibitem[{Wang et~al.(2022)Wang, Yu, Chang, Sun, and Huang}]{WangAcl2022_query}
Sijia Wang, Mo~Yu, Shiyu Chang, Lichao Sun, and Lifu Huang. 2022.
\newblock \href {https://doi.org/10.18653/v1/2022.findings-acl.16} {Query and extract: Refining event extraction as type-oriented binary decoding}.
\newblock In \emph{Findings of the Association for Computational Linguistics: ACL 2022}, pages 169--182, Dublin, Ireland. Association for Computational Linguistics.

\bibitem[{Wang et~al.(2023{\natexlab{b}})Wang, Zhou, Zu, Xia, Chen, Zhang, Zheng, Ye, Zhang, Gui, Kang, Yang, Li, and Du}]{wang2023instructuie}
Xiao Wang, Weikang Zhou, Can Zu, Han Xia, Tianze Chen, Yuansen Zhang, Rui Zheng, Junjie Ye, Qi~Zhang, Tao Gui, Jihua Kang, Jingsheng Yang, Siyuan Li, and Chunsai Du. 2023{\natexlab{b}}.
\newblock \href {https://arxiv.org/abs/2304.08085} {Instructuie: Multi-task instruction tuning for unified information extraction}.
\newblock \emph{Preprint}, arXiv:2304.08085.

\bibitem[{Wang et~al.(2023{\natexlab{c}})Wang, Li, and Ji}]{wang-etal-2023-code4struct}
Xingyao Wang, Sha Li, and Heng Ji. 2023{\natexlab{c}}.
\newblock \href {https://doi.org/10.18653/v1/2023.acl-long.202} {{C}ode4{S}truct: Code generation for few-shot event structure prediction}.
\newblock In \emph{Proceedings of the 61st Annual Meeting of the Association for Computational Linguistics (Volume 1: Long Papers)}, pages 3640--3663, Toronto, Canada. Association for Computational Linguistics.

\bibitem[{Wei et~al.(2024)Wei, Cui, Cheng, Wang, Zhang, Huang, Xie, Xu, Chen, Zhang, Jiang, and Han}]{wei2024chatie}
Xiang Wei, Xingyu Cui, Ning Cheng, Xiaobin Wang, Xin Zhang, Shen Huang, Pengjun Xie, Jinan Xu, Yufeng Chen, Meishan Zhang, Yong Jiang, and Wenjuan Han. 2024.
\newblock \href {https://arxiv.org/abs/2302.10205} {Chatie: Zero-shot information extraction via chatting with chatgpt}.
\newblock \emph{Preprint}, arXiv:2302.10205.

\bibitem[{Wu et~al.(2024)Wu, Ahmad, Zhang, Ramanathan, and Ma}]{wu2024repoformer}
Di~Wu, Wasi~Uddin Ahmad, Dejiao Zhang, Murali~Krishna Ramanathan, and Xiaofei Ma. 2024.
\newblock \href {https://arxiv.org/abs/2403.10059} {Repoformer: Selective retrieval for repository-level code completion}.
\newblock \emph{Preprint}, arXiv:2403.10059.

\bibitem[{Wu et~al.(2023)Wu, Bansal, Zhang, Wu, Li, Zhu, Jiang, Zhang, Zhang, Liu, Awadallah, White, Burger, and Wang}]{wu2023autogen}
Qingyun Wu, Gagan Bansal, Jieyu Zhang, Yiran Wu, Beibin Li, Erkang Zhu, Li~Jiang, Xiaoyun Zhang, Shaokun Zhang, Jiale Liu, Ahmed~Hassan Awadallah, Ryen~W White, Doug Burger, and Chi Wang. 2023.
\newblock \href {https://arxiv.org/abs/2308.08155} {Autogen: Enabling next-gen llm applications via multi-agent conversation}.
\newblock \emph{Preprint}, arXiv:2308.08155.

\bibitem[{Yang and Kuchibhotla(2024)}]{yang2024selection}
Yachong Yang and Arun~Kumar Kuchibhotla. 2024.
\newblock \href {https://arxiv.org/abs/2104.13871} {Selection and aggregation of conformal prediction sets}.
\newblock \emph{Preprint}, arXiv:2104.13871.

\bibitem[{Zhang et~al.(2023)Zhang, Chen, Zhang, Keung, Liu, Zan, Mao, Lou, and Chen}]{zhang2023repocoder}
Fengji Zhang, Bei Chen, Yue Zhang, Jacky Keung, Jin Liu, Daoguang Zan, Yi~Mao, Jian-Guang Lou, and Weizhu Chen. 2023.
\newblock \href {https://openreview.net/forum?id=q09vTY1Cqh} {Repocoder: Repository-level code completion through iterative retrieval and generation}.
\newblock In \emph{The 2023 Conference on Empirical Methods in Natural Language Processing}.

\bibitem[{Zhang et~al.(2022)Zhang, Roller, Goyal, Artetxe, Chen, Chen, Dewan, Diab, Li, Lin, Mihaylov, Ott, Shleifer, Shuster, Simig, Koura, Sridhar, Wang, and Zettlemoyer}]{zhang2022opt}
Susan Zhang, Stephen Roller, Naman Goyal, Mikel Artetxe, Moya Chen, Shuohui Chen, Christopher Dewan, Mona Diab, Xian Li, Xi~Victoria Lin, Todor Mihaylov, Myle Ott, Sam Shleifer, Kurt Shuster, Daniel Simig, Punit~Singh Koura, Anjali Sridhar, Tianlu Wang, and Luke Zettlemoyer. 2022.
\newblock \href {https://arxiv.org/abs/2205.01068} {Opt: Open pre-trained transformer language models}.
\newblock \emph{Preprint}, arXiv:2205.01068.

\bibitem[{Zhao et~al.(2023)Zhao, Gong, Yang, Dong, Lu, and Li}]{zhao2023demosg}
Gang Zhao, Xiaocheng Gong, Xinjie Yang, Guanting Dong, Shudong Lu, and Si~Li. 2023.
\newblock \href {https://openreview.net/forum?id=ux826WlJtt} {Demo{SG}: Demonstration-enhanced schema-guided generation for low-resource event extraction}.
\newblock In \emph{The 2023 Conference on Empirical Methods in Natural Language Processing}.

\bibitem[{Zhou et~al.(2023)Zhou, Jiang, Li, Wu, Wang, Qiu, Zhang, Chen, Wu, Wang, Zhu, Chen, Zhang, Tang, Zhang, Chen, Cui, and Sachan}]{zhou2023agents}
Wangchunshu Zhou, Yuchen~Eleanor Jiang, Long Li, Jialong Wu, Tiannan Wang, Shi Qiu, Jintian Zhang, Jing Chen, Ruipu Wu, Shuai Wang, Shiding Zhu, Jiyu Chen, Wentao Zhang, Xiangru Tang, Ningyu Zhang, Huajun Chen, Peng Cui, and Mrinmaya Sachan. 2023.
\newblock \href {https://arxiv.org/abs/2309.07870} {Agents: An open-source framework for autonomous language agents}.
\newblock \emph{Preprint}, arXiv:2309.07870.

\end{thebibliography}

\appendix

\section{Experimental Details}
\label{sec:exp_details}
The initial conformal generation risk threshold is determined by a randomly sampled calibration set from the training set. And the conformal calibration is conducted by a frozen \texttt{Flan-t5-xxl}. For ED, the initial conformal generation risk $\hat{q}_0$ is set to 1, with a decay rate $\beta$ of 0.5. For EAE, the initial conformal generation risk $\hat{q}_0$  is set to 3, also with a decay rate of 0.5. All debates are capped at a maximum of three rounds. The initial cluster radius $\mu_0$ is constantly set to 1.35, and the radius decay factor $\lambda$ is 0.9.

In our experiments, we employ three LLMs: \texttt{Llama-3-8B-Instruct} (Llama3), \texttt{Gemini-Pro} (Gemini), and \texttt{GPT-3.5-turbo} (GPT).
The Llama checkpoint is accessible at the Huggingface \cite{llama3modelcard} under Llama 3 Community License Agreement. 
We use official API to access Gemini and GPT under commercial license. Detailed description for GPT-3.5-turbo is accessible at   
\url{https://platform.openai.com/docs/models/gpt-3-5-turbo}. Detailed description for Gemini-Pro is accessible at \url{https://ai.google.dev/gemini-api/docs/get-started/tutorial?lang=python}. Additionally for the calibration model Flant5-xxl, the checkpoint is accessible at \url{https://huggingface.co/google/flan-t5-xxl} under Apache-2.0 license. No tuning is involved for any of the LLMs. All the experiments are run with one NVIDIA A40. We use Spacy for argument head detection. 

The implementation code will be made publicly available.

\section{Detailed Prompts}
\label{sec:prompts}
\paragraph{Debater Prompt for ED}
Consider the sentence: "[SENT]". Carefully read the event definition, event type, and trigger tokens in the given examples. Examine whether it mentions any possible event from the provided list. If no events are mentioned, respond with "[]". If an event are mentioned, determine the event type from the list. Then identify the event trigger, which is **one word** closely associated with the occurrence of a pre-defined event type. Respond in the format **[ROLE]: ["event type", "trigger token"]**, or **[ROLE]: []** if no event trigger is identified.

\paragraph{Debater Prompt for EAE/EE}
Give a sentence: **[SENT]**, it contains an event mention. The event type is **\{event type\}**, and the event is triggered by the token **\{trigger\}**. Now let's focus on the Argument Extraction task.
The list of argument roles corresponding to the event type **\{event type\}** is **\{role list\}**. 
Event arguments are entities that directly relate to the event mention. Please extract the event arguments of the above sentence according to the argument roles, and return them in the form of a table. 
The header of the table is | event type | argument role | argument content |.
If no entity in the sentence plays the corresponding argument role, its argument content returns **None**.

\paragraph{Critic Prompt for ED}
Review the given sentence: \"[SENT]\". Thoroughly evaluate the event definitions, typical triggers, listed examples, and responses from Debater A and Debater B. For debaters' answers, rigorously examine: Is there an event mention? Does the identified event trigger indeed express an occurrence of the identified event type, based on the event definition? Does the identified trigger align with typical triggers and the examples provided? Considering the valid examples, is there a more suitable trigger token to express the event? Provide concise assessments.

\paragraph{Critic Prompt for EAE/EE}
Remember the given sentence: **[SENT]**. Now, please judge critically and identify possible errors. Do the identified argument roles correctly match the entity mentions? Are there extra or missing argument roles, or misclassified argument roles? Please reply concisely.

\paragraph{Judge Prompt for ED}
If all agents state there is no event mention involved, reply **No event**. If all agents have agree with the same event type and event trigger answers, respond in a table. The header of the table is | event type | event trigger |. If there is any disagreement in responses, respond with **No agreement, debate continues** to encourage further discussion to resolve the differences. 

\paragraph{Judge Prompt for EAE/EE}
If debaters agree with each other, reply the event arguments in the form of a table. The header of the table is | event type | argument role | argument content |. If no argument role has a corresponding argument content, the argument content returns **None**. 
If debaters disagree on any argument content, require reply: **Disagreement observed, debate continues**.
Make sure reply only a table or **Disagreement observed, debate continues**

\end{document}